\documentclass[letterpaper]{article}
\usepackage{spconf,amsmath,graphicx}
\usepackage{multirow}
\usepackage{float}
\usepackage{subfigure}
\usepackage[flushleft]{threeparttable}
\usepackage{tabularx}

\usepackage{caption}
\graphicspath{{figs/}}

\title{Cross Attention Network for Semantic Segmentation}
\name{Mengyu Liu and Hujun Yin}
\address{School of Electrical and Electronic Engineering,
	     The University of Manchester,
	     Manchester, UK}
\begin{document}
%
\maketitle
\begin{abstract}
In this paper, we address the semantic segmentation task with a deep network that combines contextual features and spatial information. The proposed Cross Attention Network is composed of two branches and a Feature Cross Attention (FCA) module. Specifically, a shallow branch is used to preserve low-level spatial information and a deep branch is employed to extract high-level contextual features. Then the FCA module is introduced to combine these two branches. Different from most existing attention mechanisms, the FCA module obtains spatial attention map and channel attention map from two branches separately, and then fuses them. The contextual features are used to provide global contextual guidance in fused feature maps, and spatial features are used to refine localizations. The proposed network outperforms other real-time methods with improved speed on the Cityscapes and CamVid datasets with lightweight backbones, and achieves state-of-the-art performance with a deep backbone.  
\end{abstract}
\section{Introduction}
\label{sec:intro}

Semantic segmentation, which assigns class labels to image pixels, is a fundamental problem in computer vision. It has wide applications in satellite imagery analysis, medical image diagnostics, indoor scene understanding, etc. 
Significant progress has been made recently with the Fully Convolutional Network (FCN) \cite{long2015}, replacing the fully connected layers with full convolutions to maintain the resolution. 
One of the main problems in semantic segmentation is caused by the repeated downsampling layers. They are used for increasing receptive fields and features abstraction, and designed for image-level classification. However, segmentation requires pixel-level classification, which means that results have the same resolution as the input. Directly upsampling the feature maps would lead to coarse results. There are two main approaches to address this problem. One is to employ atrous (dilated) convolutions to enlarge the receptive field \cite{yu2015, chen2018deeplab}. 
The second is to utilize the U-shape architecture \cite{ronneberger2015} to hierarchically extract and recover contextual information.

Although these methods yield good performance, high computational costs are incurred due to high resolution feature maps and extra fusion computation. While hierarchical layers in these deep networks extract high-level features, spatial information is ignored. Recent research indicated that preserving spatial information with several branches helped achieve good results, as used in BiSeNet \cite{yu2018} and ICNet \cite{zhao2018icnet}. 

Based on these observations, we propose a Cross Attention Network (CANet) for semantic segmentation. It contains a shallow spatial branch, a deep context branch, and a Feature Cross Attention (FCA) module. Sufficient receptive fields are required for semantic segmentation to encode contextual information and extract abstract features, hence a deep network is adopted as the context branch. For the spatial branch, three convolutional layers are employed to preserve spatial information. Its objective is to refine the boundaries. The outputs of these two branches are fused in the FCA module. The attention blocks in FCA capture the channel-wise contextual and spatial-wise spatial information from the two branches.       

Main contributions are, (\romannumeral1) a Cross Attention Network, which consists of two branches to preserve spatial details and extract contextual features; (\romannumeral2) a Feature Cross Attention module to fuse the two branches and make the feature maps more informative both spatially and channel-wise; and (\romannumeral3) improved results obtained on two benchmarks, Cityscapes \cite{cordts2016} and CamVid \cite{brostow2008}, with fewer parameters and faster speed compared to the existing methods.             

\section{Related work}
\label{sec:Related work}

     
\textbf{Semantic segmentation.} Long \textit{et al}. proposed the FCN \cite{long2015} to take arbitrary sized input and produce corresponding segmentation map. Skipping connections were introduced to combine coarse and fine predictions to obtain denser feature maps. Chen \textit{et al.} proposed a series of segmentation networks called DeepLab \cite{chen2018deeplab, chen2017rethinking} and employed atrous convolutions to increase the field of view and maintain the spatial resolution without increasing the number of parameters. Another popular method is the U-Net architecture \cite{ronneberger2015}, composed of an encoder and a symmetric decoder. This architecture captures and recovers contextual information step by step, preserving spatial details. Jegou \textit{et al.} replaced the normal convolutional blocks in the original U-Net with dense blocks \cite{jegou2017}, to reuse features and provide deep supervision. Lin \textit{et al.} proposed a multi-path refinement network \cite{lin2017refinenet}, combining multi-scale features efficiently in a cascaded architecture.              

\textbf{Contextual and spatial information.} Contextual information is crucial for semantic segmentation due to multiple scales of objects. Enlarging the receptive fields to encode more contextual features is an effective approach. PSPNet \cite{zhao2017} performs multi-scale spatial pooling at the final feature maps to capture global features. In \cite{chen2018deeplab}, an Atrous Spatial Pyramid Pooling module was embedded at the end of the network to capture multi-scale information. Recent research showed that preserving spatial details helped achieve good results \cite{zhao2017, zhao2018icnet}. DeepLab \cite{chen2018deeplab}, PSPNet and DUC \cite{wang2018} employ atrous convolutions to control the resolution and preserve spatial information. Yu \textit{et al.} proposed an architecture combining two paths \cite{yu2018}, one shallow for spatial information and one deep for contextual information. In \cite{zhao2018icnet}, image pyramid was adopted to extract spatial features at different levels with shared bottom layers.

\textbf{Attention.} Attention is widely used for re-weighting features with high-level information in deep networks. Roy \textit{et al.} applied a concurrent attention module to semantic segmentation \cite{roy2018}, where features were squeezed along spatial and channel axes and then applied to the original feature maps to provide spatial and contextual information. EncNet \cite{zhang2018} introduces a context encoding module at the end of the network, to encode global contextual information and re-weight the extracted features for discriminative representations. In \cite{li2018}, high-level features containing category information were applied to low-level features in decoder to provide guidance. Dual Attention Network \cite{fu2018} employs two attention modules on top of the network to capture pixel relations and channel dependencies, respectively. 

\section{Methods}
\label{sec:Methods}
\subsection{Two branches}
Based on the existing methods, the two-branch architecture can encode spatial information and extract deep contextual features. In this architecture, a shallow branch is designed for preserving spatial information and a deep network is employed for capturing context. 

In the proposed CANet, the spatial branch only consists of three convolutional layers, and is applied to the original input image to preserve the resolution and encode spatial details. The first layer uses a standard convolution and the other layers employ depthwise separable convolutions \cite{sandler2018} with kernel size 3$\times$3. Each convolutional layer is followed by batch normalization \cite{ioffe2015} and ReLU \cite{krizhevsky2012}. Depthwise separable convolution factorizes a standard convolution into a depthwise convolution and a 1$\times$1 pointwise convolution, and performs as a group convolution, which splits the input to $N$ input channels (a single-channel filter is applied to each input channel). Then, pointwise convolution is used to combine these outputs linearly. The stride is 2 for all layers in the spatial branch, and the number of channels in each layer is 64, 128, 256, respectively. The resolution of the output feature map is 1/8 of the input image. The spatial branch can encode sufficient spatial information with less computational cost.
   
The context branch is used to extract high-level features. To realize this purpose, deep hierarchical networks and large receptive fields are required. The pre-trained MobileNetV2 \cite{sandler2018} is employed as the backbone of the context path. MobileNetV2 builds upon the idea of depthwise separable convolutions and can be used as a powerful feature extractor with high efficiency. In the context branch, the final convolutional layer of MobileNetV2 is discarded. The features of the last two stages are upsampled by deconvolutions and concatenated. The resolution of the final feature maps is 1/32 of the input image; affluent high-level features and contextual information can be extracted. With the two branches architecture, spatial information and contextual features can be encoded separately and then fused.    

\begin{figure}[t!]
	\centering    
	\subfigure[Feature Cross Attention module structure]{
		\includegraphics[width=0.47\textwidth]{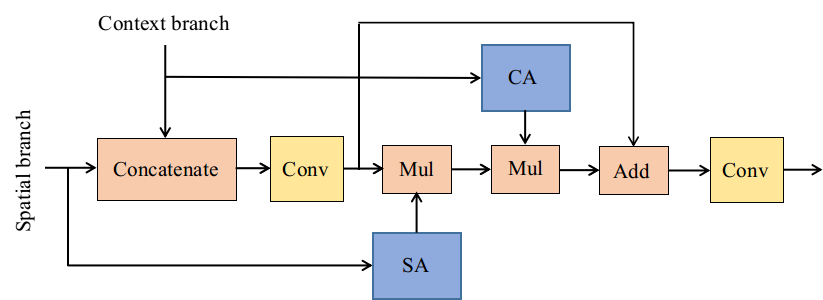}
		\label{fca}}		
	\subfigure[Spatial attention (SA) block]{
		\includegraphics[width=0.32\textwidth]{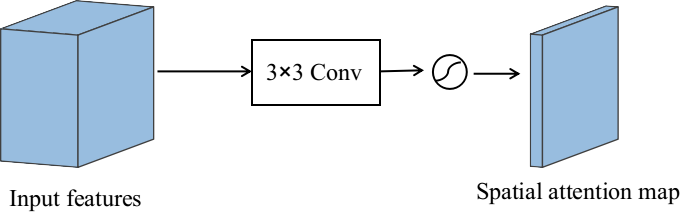}
		\label{sa}}	
	\hspace{0.1\textwidth}
	\subfigure[Channel attention (CA) block]{
		\includegraphics[width=0.46\textwidth]{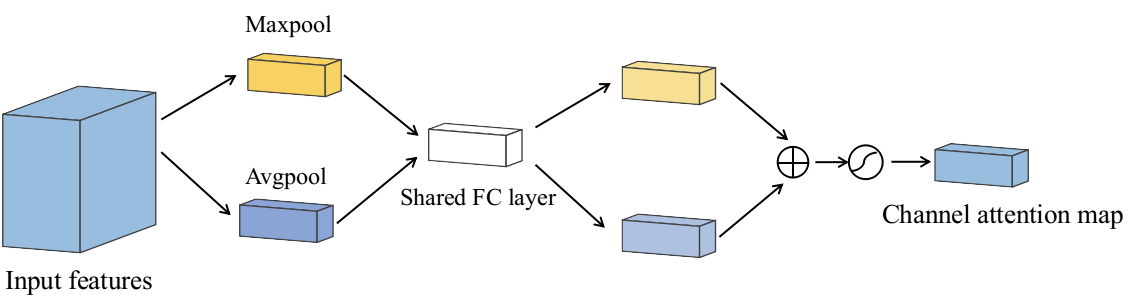}
		\label{ca}}
	\caption{Feature Cross Attention module.}
	\label{fig_sim}
\end{figure}

\subsection{Feature Cross Attention module}
Output features of the two branches are different. Features of the spatial branch contain spatial information, while features of the context branch contain contextual information. As the high-level features are mainly consists of category information, while the low-level features correspond to spatial information, it is impossible to upsample and fuse them directly. Therefore, a Feature Cross Attention (FCA) module is introduced, depicted in Fig. \ref{fca}. The high-level features from the context branch are used to provide contextual information, while the low-level features from the spatial branch are employed to refine the pixel localizations. 

In the FCA module, output features of the two branches are concatenated first, and then a 3$\times$3 convolution, batch normalization and ReLU unit are applied to the concatenated features. Next, a spatial attention block and a channel attention block are applied to the features. The spatial attention block takes the fused features and the output of the spatial branch as input, it helps refine the pixel localizations and object boundaries. Similar to the attention module in \cite{roy2018}, the features from the spatial branch go through a 3$\times$3 convolution with batch normalization and Sigmoid non-linearity, and then multiplied by the fused features. Fig. \ref{sa} shows the detail of spatial attention block, in which the 2D attention map generated from the spatial features corresponds to the importance of each pixel. It focuses on localizing the objects and refining the boundaries with the spatial information. The output of spatial attention block and contextual features from the context branch are applied to the channel attention block. The contextual features are squeezed along spatial dimensions by global pooling and max pooling to obtain two vectors. These two vectors are then applied to a shared fully connected layer and a Sigmoid operator to generate the attention map. Next the attention map is multiplied by the output features from the spatial attention block and added to the fused features. Fig. \ref{ca} presents the structure of channel attention block. The attention map reflects the importance of each channel. It focuses on the global context to provide content information. Finally, another 3$\times$3 convolution, batch normalization and ReLU unit are employed for fusion.

\subsection{Network architecture}
\begin{figure}[h!]
	\centering
	\includegraphics[scale=0.26]{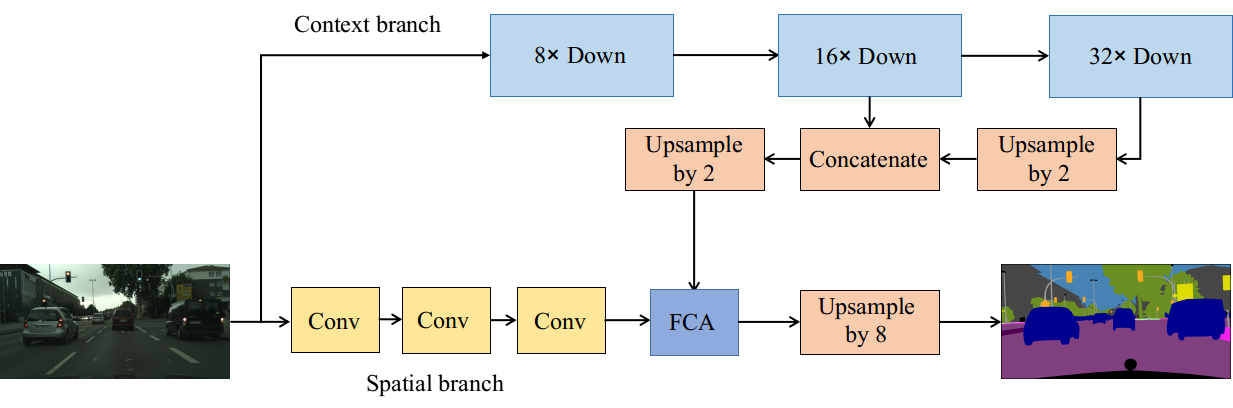}	
	\caption{Architecture of the Cross Attention Network.}
	\label{canet model}
\end{figure}

With the two-branch architecture and the FCA module, the Cross Attention Network is depicted as Fig. \ref{canet model}. We use the convolutional part of MobileNetV2 pretrained on the ImageNet in the context branch, and the final convolutional layer is discarded for reducing computational complexity. The output size of feature maps from the context branch is 1/32 of the input image. For the spatial branch, three convolutional layers are applied to the input images, the downsampling rate is 1/8. Feature maps from the last two stages are upsampled and fused by two deconvolutions. We use the FCA module to aggregate the output features from the spatial and context branches. Finally, a 1$\times$1 convolution layer is applied as the pixel-level classifier. Besides, we also use the ResNet18 and ResNet101 \cite{he2016} as backbone in the context branch. We term CANet$^1$, CANet$^2$ and CANet$^3$ to represent the CANet based on MobileNetV2, ResNet18 and ResNet101, respectively.           

\section{Experimental results}
\label{sec:Experimental results}
We evaluated our CANets on two benchmark datasets: the CamVid road scenes dataset and the urban scene dataset Cityscapes. We first introduce the implementation protocol and conduct ablation studies on the Cityscapes validation dataset, and finally we report the results on Cityscapes and CamVid datasets and compare with the state-of-the-art segmentation methods.
\subsection{Implementation protocol}     
We trained all the networks using the Adam algorithms \cite{kingma2014} with batch size 8 and weight decay 0.0001. We applied the ``ploy'' learning rate policy \cite{chen2018deeplab} where the current learning rate equalled $init\;lr \times (1-\frac{epoch}{max\_epoch})^{0.9}$, and the initial learning rate was 0.0001. We adopted random horizontal flipping and random scaling between 0.5 and 2 for data augmentation. All images were normalized to zero mean and unit variance. Weighted cross-entropy loss was used to optimize the networks due to class unbalance. Results are reported using the mean Intersection over Union (mIoU) metric:
\begin{equation}
IoU = \frac{TP}{TP+FP+FN}
\label{miou}
\end{equation}   
where TP, FP and FN are the number of true positives, false positives and false negatives at pixel-level. 

\subsection{Ablation study} 
The Cityscapes is a urban street scene dataset for semantic understanding. It contains 5000 high quality annotated images collected from 50 different cities, divided into three sets, 2975 for training, 500 for validation and 1525 for test. All images are of 2018$\times$1024, and all pixels are annotated to 19 classes. In the ablation studies, we used the validation set to investigate the effect of the FCA module, and all the input images for training were randomly cropped 713$\times$713 subimages.  

\textbf{Baseline:} We used a simple two-branch network as the baseline, where the pre-trained MobileNetV2 was used as the backbone of the context path and three convolutional layers as the spatial branch. The output features of these two branches were concatenated and passed to a 1$\times$1 convolutional layer for final classification. Evaluated performance of this baseline is shown in Table \ref{FCA}.

\textbf{Ablation for fusion module:} We applied two consecutive 3$\times$3 convolutional layers with batch normalization and ReLU units to the concatenated features to fuse them. This improved the performance from 67.9\% to 69.2\%.  

\textbf{Ablation for cross attention blocks:} First we evaluated the performance with only the spatial attention block, and then arranged the spatial and channel attention blocks in sequential or parallel order. The results are shown in Table \ref{FCA}, showing that two attention blocks boosted the performance, while the spatial-channel order achieved the best result. 

\begin{table}[t!]
	\caption{Detailed performance of different feature fusion modules on the Cityscapes validation set.} 
	\label{FCA}
	\centering
	\begin{threeparttable}	
	\begin{tabularx}{8.5cm}{p{6.1cm} | X<{\centering}}
		\hline
		Method                                    & mIoU (\%)      \\ \hline\hline
		Baseline                                  & 67.9          \\
		Baseline + C33$^{\dagger}$                & 69.2          \\
		Baseline + spatial                        & 72.3          \\
		Baseline + channel \& spatial in parallel & 73.2          \\
		Baseline + channel + spatial              & 73.1          \\
		Baseline + spatial + channel              & \textbf{73.4} \\ \hline
	\end{tabularx}
	\begin{tablenotes}
		\footnotesize
		\item $^{\dagger}$: ``C33'' denotes using two consecutive 3$\times$3 convolutional blocks as the fusion module.
	\end{tablenotes}
\end{threeparttable}
\end{table}

\begin{table}[t!]
	\caption{Speed and accuracy comparison of CANets against other networks.} 
	\label{speed}
	\centering	
	\begin{tabularx}{8.5cm}{p{2.15cm} | p{1.2cm}<{\centering} | p{0.75cm}<{\centering} | p{1.1cm}<{\centering} | X<{\centering}}
		\hline
		Method                           & FLOPs & FPS   & \#Params & mIoU(\%) \\ \hline\hline
		ICNet \cite{zhao2018icnet}       & 29.6G   & 74.4  & 7.8M   & 69.5      \\
		ERFNet \cite{romera2018}         & 25.8G   & 73.5  & 2.1M   & 69.7      \\
		DeepLabv2 \cite{chen2018deeplab} & 362.5G  & 10.9  & 43.3M  & 70.4      \\ \hline\hline
		CANet$^1$                        & 18.5G   & 95.3  & 4.8M   & 69.5      \\
		CANet$^2$                        & 38.7G   & 104.8 & 15.8M  & 70.9      \\ \hline
	\end{tabularx}
	
\end{table}

\subsection{Cityscapes} 
Based on the ablation studies, we designed the complete network architecture and experimented it on the Cityscape dataset. Most deep networks for semantic segmentation require large amounts of computational resources and run slowly even on modern GPUs. Computational speed and memory usage are important factors of a method. First, we conducted our experiments to test the inference speed in comparison with other methods. We chose downsampled 1024$\times$512 as the input size. All experiments were conducted on one NVIDIA TITAN V GPU, using PyTorch framework \cite{paszke2017} with CUDA 10.0, and each network was randomly initialized and evaluated for 100 times. The results are shown in Table \ref{speed}. Next, we trained and evaluated CANet$^1$ and CANet$^2$ at 1024$\times$512 and accuracies on the test set are shown in Table \ref{speed}. The results show that CANets outperformed other real-time methods with faster speed on the Cityscape dataset.

\begin{figure}[t!]
	\centering    
	\subfigure[Image]{
		\begin{minipage}[t]{0.145\textwidth}
			\centering
			\includegraphics[width=1\textwidth]{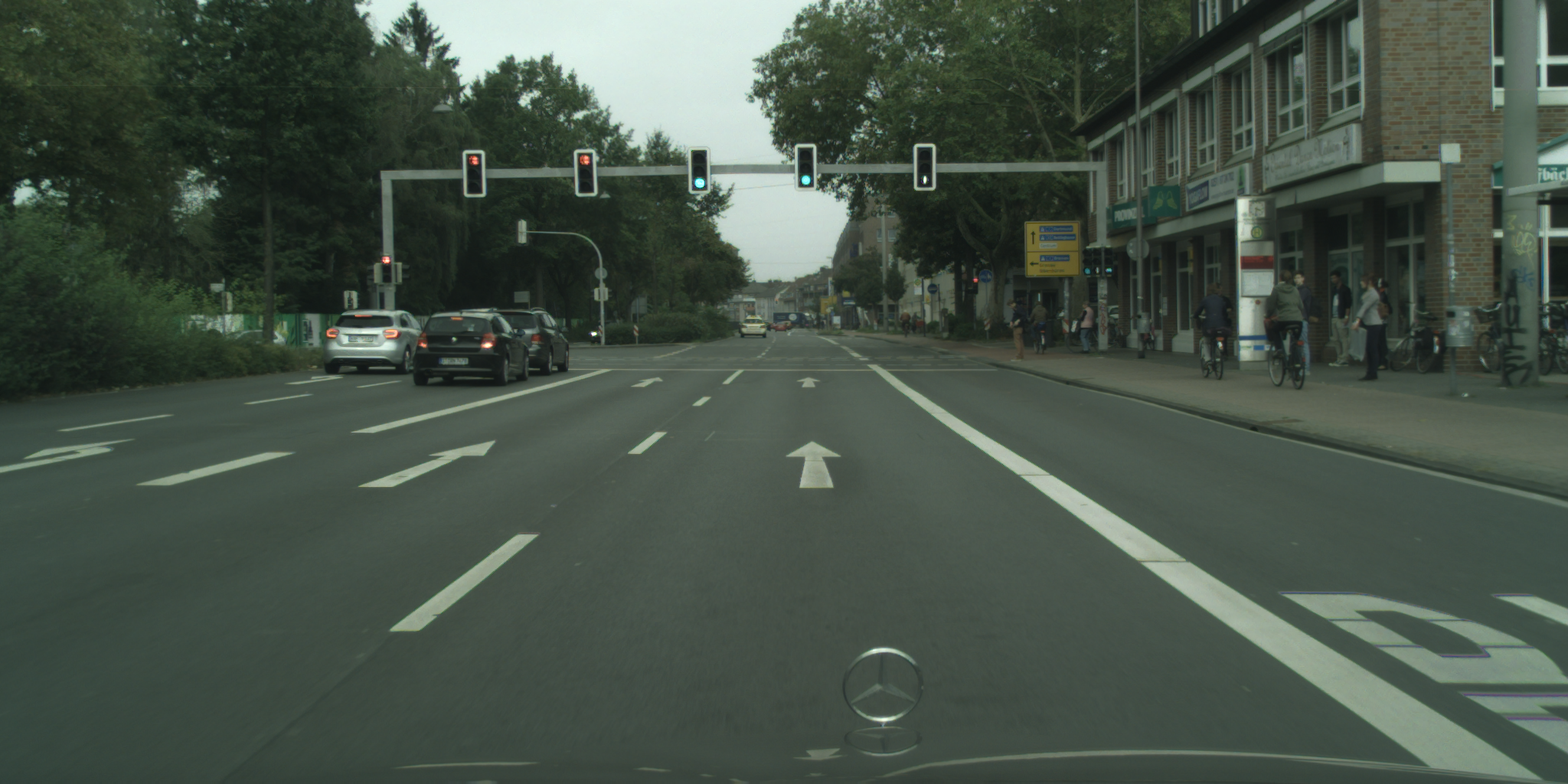}\\
			\vspace{0.02\textwidth}
			\includegraphics[width=1\textwidth]{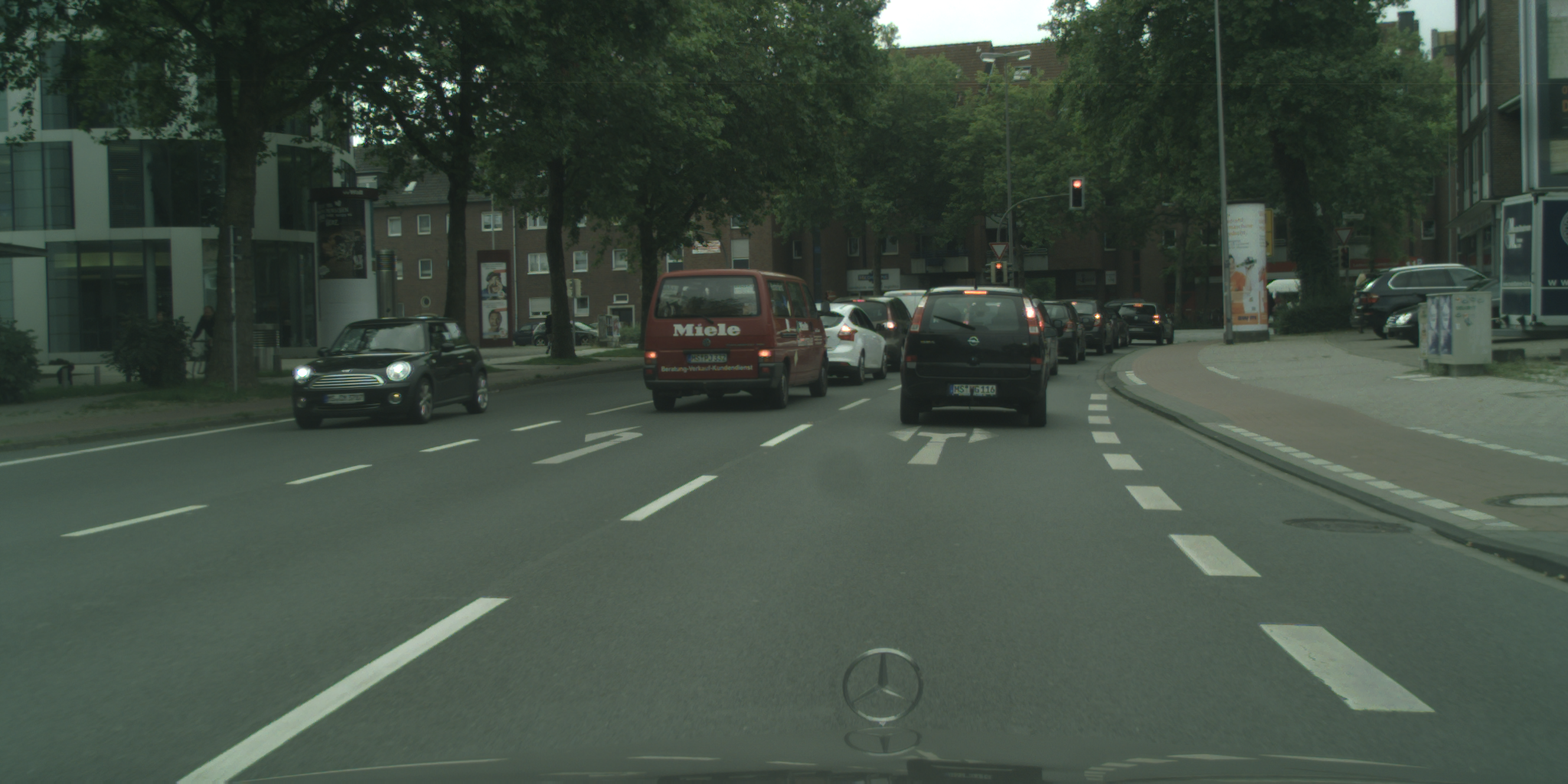}\\
			\vspace{0.2cm} 
			\label{image}	
		\end{minipage}
	}		
	\subfigure[CANet$^{3}$]{
		\begin{minipage}[t]{0.145\textwidth}
			\centering
			\includegraphics[width=1\textwidth]{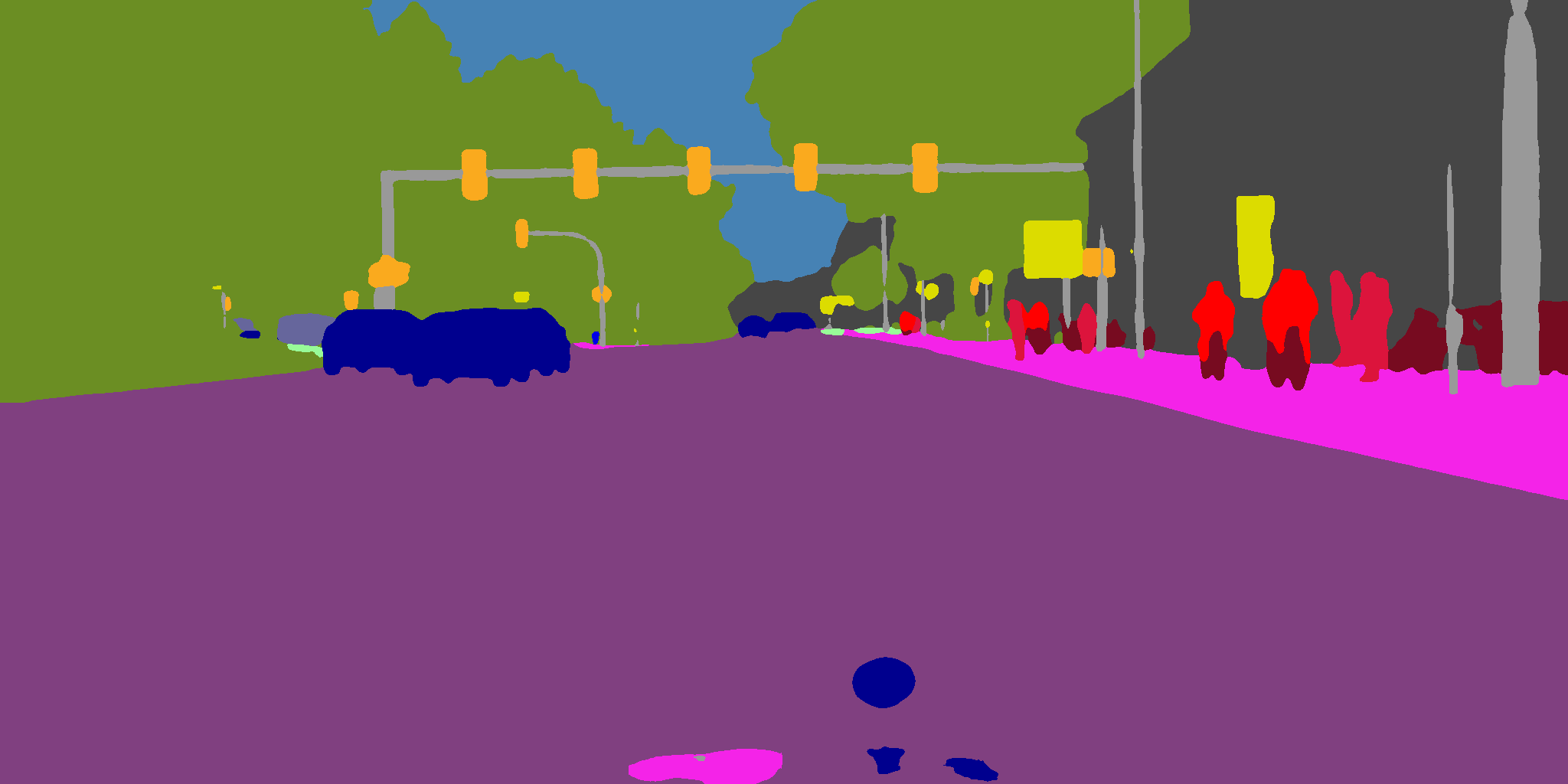}\\
			\vspace{0.02\textwidth}
			\includegraphics[width=1\textwidth]{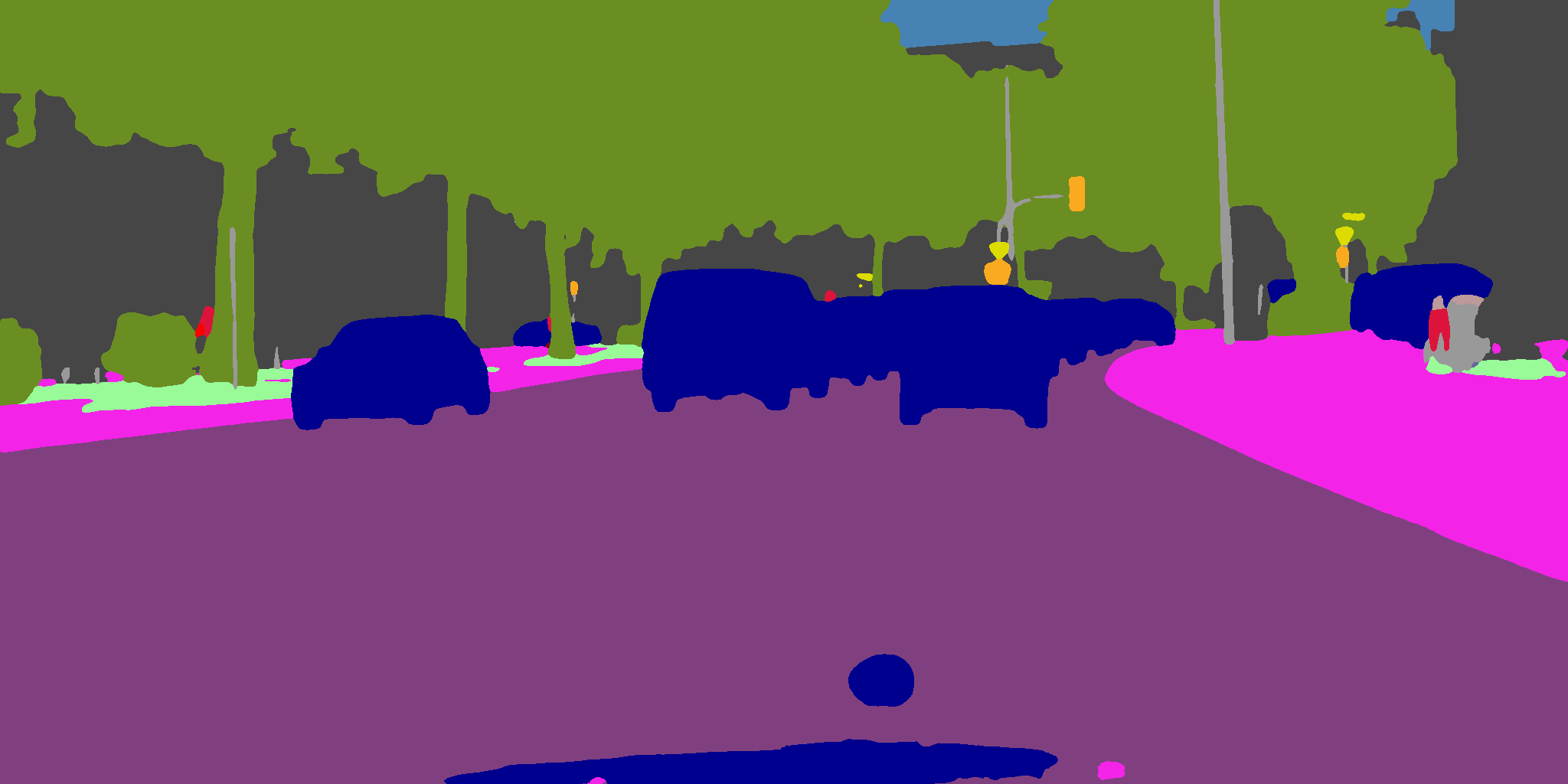}\\
			\vspace{0.2cm} 
			\label{canet}
		\end{minipage}
	}	
	\subfigure[Groundtruth]{
		\begin{minipage}[t]{0.145\textwidth}
			\centering
			\includegraphics[width=1\textwidth]{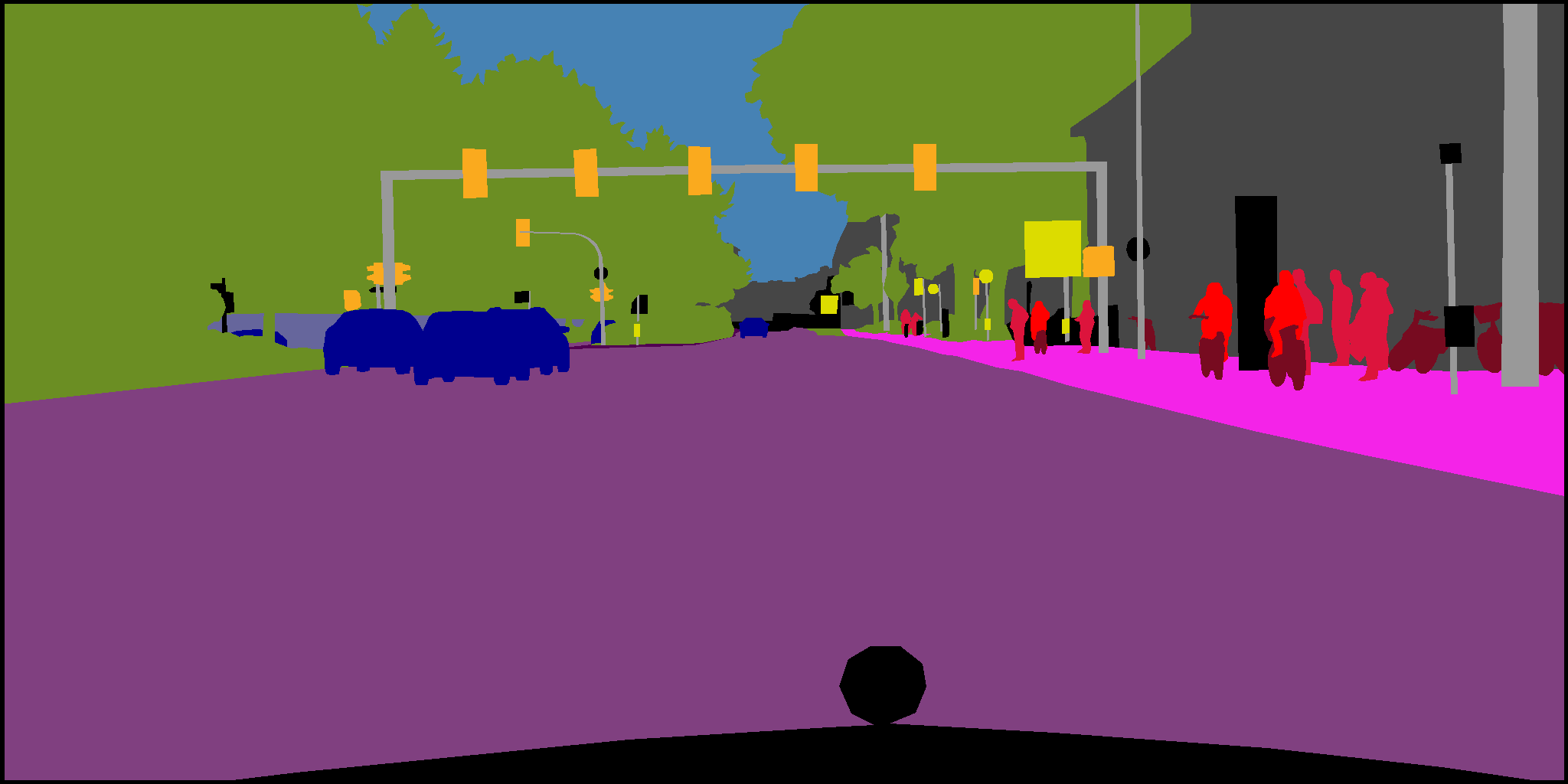}\\
			\vspace{0.02\textwidth}
			\includegraphics[width=1\textwidth]{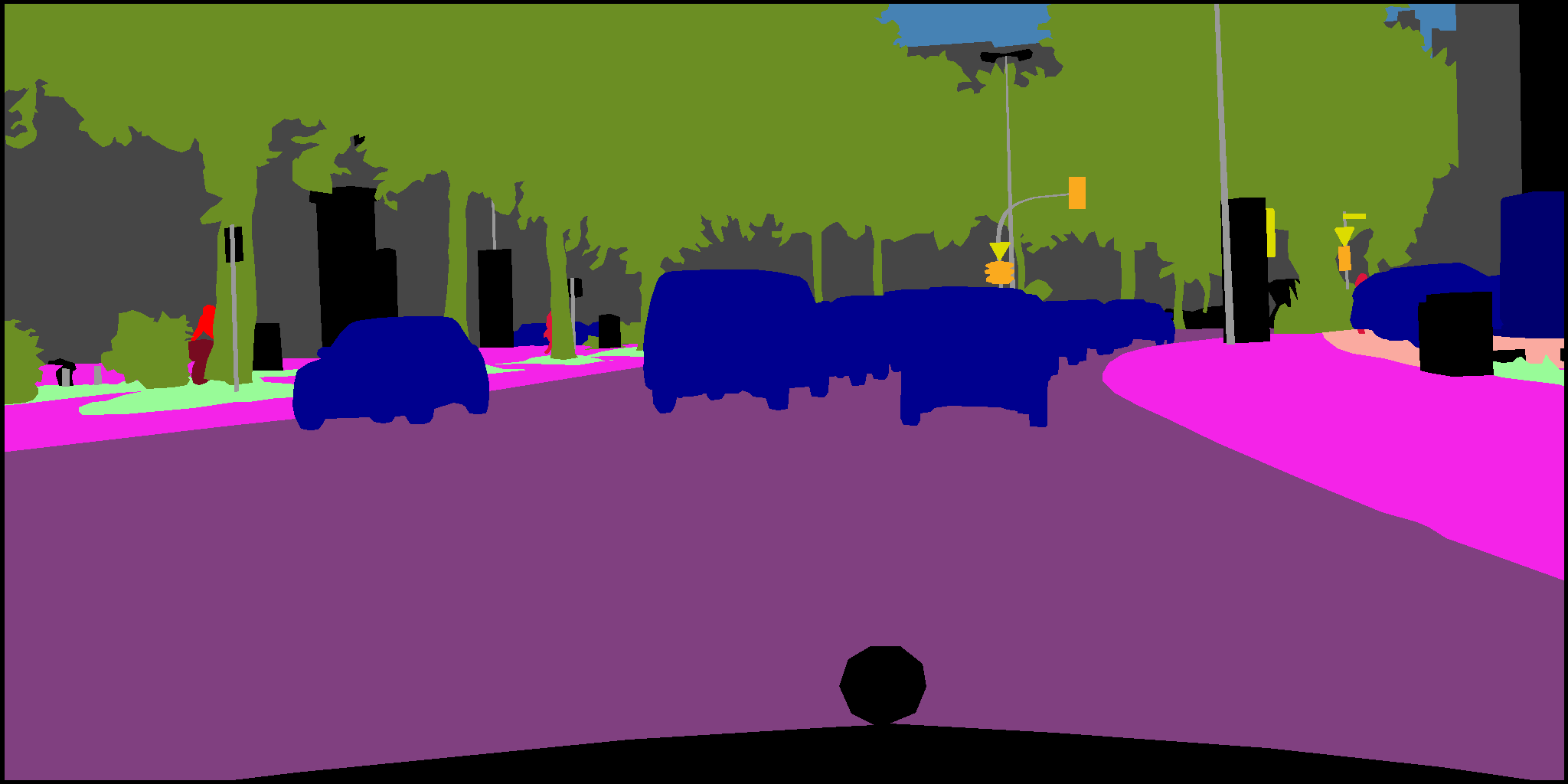}\\
			\vspace{0.2cm} 
			\label{gt}
		\end{minipage}
	}
	\caption{Visualization results on the Cityscapes dataset.}
	\label{example}
\end{figure}

Next, we trained our three networks on Cityscapes and randomly took 769$\times$769 crop as input. The results on the Cityscapes test set are summarized in Table \ref{acc}. Some visualization results of the CANet$^3$ are presented in Fig. \ref{example}. 

\begin{table}[t!]
	\caption{Results on the Cityscapes test set.} 
	\label{acc}
	\centering	
	\begin{tabularx}{8.5cm}{p{3.2cm} | p{2.3cm}<{\centering} | X<{\centering}}
		\hline
		Method                            & Backbone    & mIoU (\%) \\ \hline\hline
		PEARL \cite{jin2017}              & ResNet101   & 73.4      \\
		RefineNet \cite{lin2017refinenet} & ResNet101   & 73.6      \\
		SAC \cite{zhang2017scale}         & ResNet101   & 78.1      \\
		PSPNet \cite{zhao2017}            & ResNet101   & 78.4      \\
		DUC \cite{wang2018}               & ResNet152   & 77.6      \\
		DepthSeg \cite{kong2018recurrent} & ResNet101   & 78.2      \\ \hline\hline
		CANet$^1$                         & MobileNetV2 & 73.5      \\
		CANet$^2$                         & ResNet18    & 75.1      \\
		CANet$^3$                         & ResNet101   & 78.6      \\ \hline
	\end{tabularx}
	
\end{table}

\subsection{CamVid}
The CamVid road scenes dataset \cite{brostow2008} has fully labelled images for semantic segmentation: 367 for training, 101 for validation and 233 for test. Each image is of 480$\times$360 and labelled with 11 semantic classes. We used the MobilenetV2 and ResNet18 as the backbone of the context branch respectively, and detailed results are shown in Table \ref{camvid}.

\begin{table}[t!]
	\caption{Results on the CamVid test set.} 
	\label{camvid}	
	\centering	
	\begin{tabularx}{8.5cm}{p{3.2cm} | p{2.3cm}<{\centering} | X<{\centering}}
		\hline
		Method                                     & Global avg. (\%) & mIoU (\%) \\ \hline\hline
		FCN8 \cite{long2015}                       & 83.1             & 52.0      \\
		Bayesian SegNet \cite{kendall2015bayesian} & 86.9             & 63.1      \\
		Dilation8 \cite{yu2015}                    & 79.0             & 65.3      \\
		Dilation8 + FSO \cite{kundu2016}           & 88.3             & 66.1      \\ \hline\hline
		CANet$^1$                                  & 90.8             & 66.6      \\
		CANet$^2$                                  & 90.8             & 66.9      \\
		CANet$^3$                                  & 90.6             & 67.4     \\\hline
	\end{tabularx}
	
\end{table} 

\section{Conclusions}
This paper presents a new Cross Attention Network (CANet) for semantic segmentation. We design a two-branch network to extract high-level contextual features and encode low-level spatial information simultaneously. In the context branch, lightweight networks are employed to reduce computational cost, 
a Feature Cross Attention (FCA) module is proposed to fuse these two kinds of features. Contextual features are used to provide global information for fused features, and spatial features are employed to refine pixel localizations and object boundaries. The ablation experiments show that FCA module combines contextual features and spatial information efficiently and gives more precise segmentation. Experiment results on the Cityscapes and CamVid datasets show that the CANets outperform other real-time methods with faster speed, and achieve comparable performance to the state-of-the-art methods when employing a deep backbone.



\bibliographystyle{IEEEbib}
\bibliography{strings,refs}

\end{document}